\documentclass[conference]{IEEEtran}
\IEEEoverridecommandlockouts
\usepackage{cite}
\usepackage{tikz}
\usepackage{graphicx}
\usepackage{subcaption}
\usetikzlibrary{arrows, decorations.markings}
\usepackage{times}
\usepackage{threeparttable}
\usepackage{amsmath,amssymb,amsfonts}
\usepackage{algorithmic}
\usepackage[bookmarks=false]{hyperref}
\usepackage{graphicx}
\usepackage{algorithm}
\usepackage{algorithmic}
\usepackage{textcomp}
\usepackage{url}
\usepackage{xcolor}
\usepackage{multirow}
\usepackage{multicol}
\def\BibTeX{{\rm B\kern-.05em{\sc i\kern-.025em b}\kern-.08em
    T\kern-.1667em\lower.7ex\hbox{E}\kern-.125emX}}
\usepackage[export]{adjustbox}
\usepackage{array}
\usepackage{dblfloatfix}	
\usepackage{mathtools}
\usepackage{placeins}
\usepackage{makecell}
\usepackage{xspace}
\usepackage{wrapfig}
\usepackage{float}

\makeatletter
\def\@oddfoot{\hbox{IEEE ICDM Workshop on Data Mining for Services 2018}\hfil \scriptsize \thepage}
\def\@evenfoot{\hbox{}\hfil \scriptsize \thepage}
\makeatother

\begin{document}

\title{Generating Realistic Sequences of Customer-level Transactions for Retail Datasets\\
}

\author{\IEEEauthorblockN{Thang Doan$^*$ \thanks{$^*$ Work done as an intern at Rubikloud}}
\IEEEauthorblockA{\textit{McGill University} \\
Montreal, Canada \\
thang.doan@mail.mcgill.ca}
\and
\IEEEauthorblockN{Neil Veira$^*$}
\IEEEauthorblockA{\textit{University of Toronto} \\
Toronto, Canada \\
nveira@eecg.toronto.edu}

\\
\IEEEauthorblockN{Brian Keng}
\IEEEauthorblockA{\textit{Rubikloud Technologies Inc.} \\
Toronto, Canada \\
brian.keng@rubikloud.com}

\and
\IEEEauthorblockN{Saibal Ray}
\IEEEauthorblockA{\textit{McGill University} \\
Montreal, Canada \\
saibal.ray@mcgill.ca}
}

\maketitle

\begin{abstract}
In order to better engage with customers, retailers rely on extensive customer
and product databases which allows them to better understand customer behaviour
and purchasing patterns.  This has long been a challenging task as customer
modelling is a multi-faceted, noisy and time-dependent problem.  The most
common way to tackle this problem is indirectly through task-specific
supervised learning prediction problems, with relatively little literature on
modelling a customer by directly simulating their future transactions.
In this paper we propose a method for generating realistic
sequences of baskets that a given customer is likely to purchase over a period
of time.
Customer embedding representations are learned using a Recurrent Neural Network
(RNN) which takes into account the entire sequence of transaction data.
Given the customer state at a specific point in time, a Generative Adversarial
Network (GAN) is trained to generate a plausible basket of products for the
following week.  The newly generated basket is then fed back into the RNN to
update the customer's state.  The GAN is thus used in tandem with the RNN
module in a pipeline alternating between basket generation and customer state
updating steps.  This allows for sampling over a distribution of a customer's
future sequence of baskets, which then can be used to gain insight into how to
service the customer more effectively.  The methodology is empirically shown to
produce baskets that appear similar to real baskets and enjoy many common
properties, including frequencies of different product types, brands, and
prices.  Furthermore, the generated data is able to replicate most of the strongest
sequential patterns that exist between product types in the real data.

\end{abstract}


\section{Introduction}
Modern retailers collect, store and utilize massive amounts of consumer
behaviour data through their customer loyalty programs.  Sources such as
customer-level transactional data, customer profiles, and product attributes
allow the retailer to better service their customers by utilizing data mining
techniques for customer relationship management (CRM) and direct marketing
systems~\cite{LinoffBerry2004}.  Better data mining techniques for CRM databases can allow retailers
to understand their customers more effectively, leading to increased loyalty, 
better service and ultimately increased sales.

Modelling customers is a complex problem with many facets.  First, a retailer's
loyalty data provides a censored view into a customer's behaviour because it
only shows the transactions for a single retailer, leading to noisy observations.
Second, the sequential nature of consumer purchases adds additional complexity
as changes in behaviour and long term dependencies need to be taken into
account.  Finally, the large number of customers (100M+) multiplied by the
catalog of products (100K+) results in a vast amount of transactional data, but it is simultaneously
very sparse at the level of individual customers.  These complexities
make modelling customers a difficult problem even with modern techniques.

Indirect approaches to modelling customer behaviour for specific tasks have been
widely studied.  Techniques that utilize customer-level transactional data 
such as customer lifetime value~\cite{customer_lifetime_value}, recommendations~\cite{LSTM_basket,collaborative_recommender}, 
and incremental sales~\cite{Radcliffe2011a},
formulate these tasks as supervised learning problems.  
More direct approaches to modelling customers
have been through simulators.  There are a wide variety of applications for customer
marketing simulators from aiding in decision support~\cite{SchwaigerStahmer2003} to understanding how
behavioural phenomena affect consumer decisions~\cite{ZhangZhang2007}.
Another notable application of customer simulators is in the context of
direct marketing activities~\cite{Tkachenko2016,Abe2004,Pednault2002,Silver2013}.
These methods use the customer simulator to understand individual-level
interactions between a customers and a marketing agent.  Typically, the goal is
to find an ideal marketing policy to maximize a pre-defined reward over time.
However, the primary focus of this work has been on techniques for
generating an optimal marketing policy with less focus on generating realistic
simulations of customer transactional data.

Generative modelling methods~\cite{GAN} have proven to be very successful in 
learning realistic distributions from the data in many different contexts.  
A relevant recent work by~\cite{EcommerceGAN}
presents a technique to generate realistic orders from an e-commerce dataset.
They provide a method to effectively learn the complex relationships between
customer and product to generate realistic simulations of customer orders,
but do not take into account how customer behaviour changes over time in their method.

In this work, we present a novel method to generate realistic sequences of customer-level baskets of products over time using a customer-level retail transactions dataset. Our technique is able to generate
samples of both customers and traces of their transaction baskets over time.  
This general formulation of the customer modelling problem allows one to
essentially generate new customer-level transactional datasets that retain most of the
distributional properties of the original data.
This opens up possibilities for new applications such as generating a distribution of
likely products to be purchased by an individual customer in the future to derive insights for better service, 
or by providing external researchers with access to generated 
data for a source dataset that otherwise would be restricted due to privacy concerns.

The proposed method uses a multi-step approach to generating customer-level 
transactional data using using a combination of Generative Adversarial Networks
(GAN) \cite{GAN} and Recurrent Neural Networks (RNN) \cite{Hochreiter1997}.
First, we train a RNN to generate a customer embedding by using a multi-task
learning approach.  The inputs to the RNN are product embeddings derived from
their textual descriptions.  This allows one to describe the customer state
given their previous transactions.
Next, to determine the number of products in the next basket, we extract a sample based on historical basket sizes of similar customers.
A GAN trained by conditioning on a customer embedding at the current time is
used to predict the next product in a basket for a given customer.  This is
repeated until all products in the basket are filled.  This provides a single
customer-level transaction basket.
Finally, the generated products are fed back into the RNN to generate the next
state of the customer and the process repeats.

Evaluation of GANs and generative models are difficult in general~\cite{Theis2015,Zhao2016} especially for non-visual domains.  We demonstrate the effectiveness of the
technique via several qualitative and quantitative metrics.
We first show that the generator can reproduce the relative frequencies of various product features including types, brands, and prices to within a 5\% difference. 
We further show that the generated data retains most of the strongest sequential patterns between products in the real data set.
Finally, we show that most of the real and generated baskets are indistinguishable, with a classifier trained to separate the two being able to achieve an accuracy of only 63\% at the category level.


\section{Background and Related Work} \label{sec:related}
\subsection{Transaction-Based Item and Customer Embeddings}
Learning a representation of customers from their transactional data is a
common problem in retail data mining. 
Borrowing inspiration from Natural Language Processing (NLP), 
different methods try to embed customers into a common vector space based
of their transaction sequences. For instance, \cite{trans2vec} and
\cite{item2vec} learn the embeddings by adapting the Paragraph
Vector-Distributed Bag-of-Words or the n-skip-gram models from~\cite{word2vec}.
The underlying idea behind these methods is that by solving an intermediate
task such as predicting the next word in a sentence or the next item a customer
will purchase, one can learn general--purpose features that are meaningful and have good predictive power for a wide variety of tasks. 

For example, \cite{customer_lifetime_value} examines the lifetime
value of a customer in the context of an e-commerce website.  Towards that end,
they also use an n-skip-gram model to learn customer embeddings and track its
evolution over time as purchases are made.  
\cite{client2vec} uses a stacked denoising autoencoder to learn customer embeddings for improving
their campaign decisions or clustering clients into classes. 


\subsection{Item Prediction and Recommendation Systems}
Various techniques from recommendation
systems such as collaborative
filtering~\cite{collaborative,collaborative_recommender} have long been used to
predict a customer's preference for items, although usually they are not
directly predicting a customer's next purchase.

More recent advancements in deep learning have shown to be quite practical 
in modelling a customer's next purchase over time.  Techniques such
as~\cite{next_basket_with_neural_nets} mimic a recurrent neural network (RNN)
by feeding historical transaction data as input to a
neural network which predicts the next item.  \cite{LSTM_basket} and
\cite{attribute_aware} both use a RNN to predict the next basket of items to
great effect.


\subsection{Generative Adversarial Networks}
Generative Adversarial Networks (GANs)~\cite{GAN} are a class of generative models aimed at learning a distribution. 
The method is founded on the game theoretical concept of two-player zero-sum games, wherein two players each try to maximize their own utility at the expense of the other player's utility. 
By formulating the distribution learning problem as such a game, a GAN can be trained to learn good strategies for each player.
A generator $G$ aims to produce realistic samples from this distribution while a discriminator $D$ tries to differentiate fake samples from real samples. 
By alternating optimization steps between the two components, the generator ultimately learns the distribution of the real data.

In detail, the generator network $G:Z \to X$ is a mapping from a high-dimensional noise space $Z = \mathbb{R}^{d_z}$ to the input space $X$ on which a target distribution $f_X$ is defined. The generator's task consists of fitting the underlying distribution of observed data $f_X$ as closely as possible. The discriminator network $D:X \to \mathbb{R} \cap [0,1]$ scores each input with the probability that it belongs to the real data distribution $f_X$ rather than the generator $G$.

The classical GAN optimization algorithm minimizes the Jensen-Shannon divergence (JS) between the real and generated distributions.
However, \cite{wgan} suggests replacing the JS metric by the Wasserstein-1 or Earth-Mover divergence. We make use of an improved version of this algorithm, the Wasserstein GAN (WGAN) with Gradient Penalty \cite{wgan-gp}. Its objective is given below:
\begin{equation}
    \min_G\max_D \underset{{x \sim f_X(x)}}{\mathbb{E}}[D(x)]+\underset{{x \sim G(z)}}{\mathbb{E}}[-D(x)]+p(\lambda),
    \label{eq:wass_gan_loss}
\end{equation}
where $p(\lambda)=\lambda(||\nabla_{\tilde{x}} D(\tilde{x})||-1)^2,\; \tilde{x}=\varepsilon x + (1-\varepsilon)G(Z)$,  $\varepsilon \sim \text{Uniform}(0,1)$, and $Z\sim f_Z(z)$. Setting $\lambda=0$ recovers the original WGAN objective.\\

\subsection{Simulating Customer Behaviour}
A customer's state with respect to a given retailer (\textit{i.e.} the types of products they are
interested in and the amount they are willing to spend) evolves over
time, and there exist a wide variety of techniques used to model this state.
In marketing research, agent-based approaches such as~\cite{SchwaigerStahmer2003,ZhangZhang2007} 
have aided in building simple simulations of how customers interact and make decisions.

Data mining and machine learning approaches to model a customer's state in the
context of direct marketing activities have also been widely
studied~\cite{Tkachenko2016}.  Techniques such
as~\cite{Abe2004,Pednault2002,Silver2013} model the problem in the
reinforcement learning framework by attempting to learn the optimal marketing
policy to maximize rewards over time.  As part of their work, they use various
techniques to represent and simulate the customer's state over time. 
However, the method does not use the customer's state to generate its future orders,
but rather consider it more narrowly in the context of the defined reward.

More recently,~\cite{EcommerceGAN} was the first to generate plausible customer
e-commerce orders for a given product using a Generative Adversarial Network (GAN). 
Given a product embedding, \cite{EcommerceGAN} generates a tuple containing a
product embedding, customer embedding, price, and date of purchases,
which summarizes a typical order.  The e-commerce GAN has applications
in providing insights into product demand, customer preferences, price
estimation and seasonal variations by simulating what are likely potential
orders.  However, it only generates realistic orders and does not directly model customer
behaviour over time.

\section{Methodology} \label{sec:methodology}
In this section, we present a novel methodology for generating realistic sequences of future transactions for a given customer. 
The proposed pipeline involves a GAN module and an LSTM module intertwined in a sequence of product generation and customer state updating steps. 
The GAN is trained to generate a basket of products conditioned on a time-sensitive customer representation, while the LSTM models the sequential nature of the customer's state as it interacts with products. 
Each of these components uses semantic embeddings of customers and products for representational purposes, which are defined in the first two subsections, while the training of the GAN and generation of customer transactions are presented in the latter two subsections.

\subsection{Product Representations} \label{sec:product_rep}

To capture the semantic relationships between products that exist independently of customer interactions, we learn product representations based on their associated textual descriptions. 
Specifically, a corpus is created wherein a sentence is defined for each product as the concatenation of the product name and description.
Preprocessing is applied to remove stopwords and other irrelevant tokens. 
The resulting corpus contains $11,443$ products and a vocabulary size of $21,894$ words.
The word2vec skipgram model~\cite{word2vec} is then trained on this corpus using a context window size of 5 and an embedding dimensionality of 128.
Finally, each product representation is defined as the arithmetic mean of the word embeddings in the product's name and description.
This is similar to the common practice of representing a sentence by the mean of word vectors within the sentence~\cite{le2014distributed}, and is motivated by the observation that sums of word vectors produce semantically meaningful vectors.

\subsection{Customer Representations} \label{sec:customer_rep}
To characterize customers by their purchasing habits we learn customer embedding representations from their transactional data. 
Inspired by~\cite{EcommerceGAN}, this is accomplished using a Long-Short Term Memory (LSTM) \cite{LSTM} module.
The LSTM takes as input a sequence of transaction baskets for a given customer, where each transaction basket is defined by a set of product embeddings for a week of purchase.
Products within the same basket are ordered randomly during training.
The LSTM is trained to learn the customer's sequential patterns via a multi-task optimization procedure. Specifically, the LSTM output is fed as inputs for the following three prediction tasks:
\begin{enumerate}
    \item Predict whether or not a product is the last product in the basket. 
    \item Predict the category of the next product.
    \item Predict the price of the next product.
\end{enumerate}

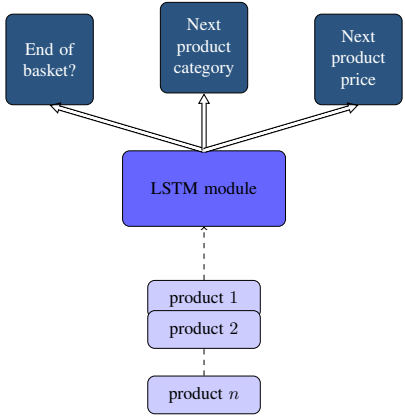
\begin{figure}
\centering
\resizebox{0.3\textwidth}{!}{
\begin{tikzpicture}[node distance=0.8cm][transform canvas={scale=0.4}]

\tikzstyle{LSTM} = [rectangle,rounded corners, minimum width=3cm, minimum height=1.5cm, text centered,text width=3cm, draw=black, fill=blue!60]
\tikzstyle{product} = [rectangle, rounded corners, minimum width=2cm, minimum height=0.75cm,text centered,text width=2cm, draw=black, fill=blue!20]
\tikzstyle{TASK} = [rectangle,rounded corners, minimum width=1.5cm, minimum height=1.8cm, text centered,text width=1.5cm, draw=black, fill={rgb:red,1;green,2;blue,3}]
\tikzstyle{arrow} = [draw, -latex']
\tikzstyle{vecArrow} = [thick, decoration={markings,mark=at position
   1 with {\arrow[semithick]{open triangle 60}}},
   double distance=1.4pt, shorten >= 5.5pt,
   preaction = {decorate},
   postaction = {draw,line width=1.4pt, white,shorten >= 4.5pt}]
\tikzstyle{innerWhite} = [semithick, white,line width=1.4pt, shorten >= 4.5pt]

\node (lstm) [LSTM]      {LSTM module};
\node (task1)[TASK,above left of=lstm,yshift=2.0cm, xshift=-2.5cm]{\textcolor{white}{End of basket?}};
\node (task2)[TASK,above of=lstm,yshift=2.0cm] {\textcolor{white}{Next product category}};
\node (task3)[TASK,above right of=lstm,yshift=2.0cm,xshift=2.5cm] {\textcolor{white}{Next product price}};

\node (product1) [product, below of=lstm, yshift=-1.40cm]  {product $1$};
\node (product2) [product, below of=product1, yshift=+0.20cm]  {product $2$};
\node (productn) [product, below of=product2, yshift=-0.5cm] {product $n$};

\draw[dashed,->]  (product1.north) -- (lstm.south) ;
\draw[dashed] (productn) -- (product2);

\draw[vecArrow] (lstm.north) to (task1.south);
\draw[innerWhite] (lstm.north) to (task1.south);

\draw[vecArrow] (lstm.north) to (task2.south);
\draw[innerWhite] (lstm.north) to (task2.south);

\draw[vecArrow] (lstm.north) to (task3.south);
\draw[innerWhite] (lstm.north) to (task3.south);


\end{tikzpicture}
}
\caption{Embedding customers via multi-task Learning with an LSTMs.  The input is the sequence of
products a customer has purchased throughout their transactional history.  After convergence, the hidden
state of the LSTM will characterize a customer's state.}
\label{fig:tikz_multi_task}
\end{figure}

The LSTM is trained to maximize the performance of all three subtasks by randomly and uniformly sampling a single task in each step and optimizing for this task.
After convergence, the hidden state of the LSTM is used to characterize a customer's purchasing habits,
and thus a customer's state.
As a result, customers with similar behaviour will be closer together in the resulting embedding space. 
Figure \ref{fig:tikz_multi_task} illustrates the process of learning this embedding. 

\subsection{Learning Product Distributions with a Conditional GAN}  \label{sec:GAN}
To learn the product distributions, we use a conditional Wasserstein GAN~\cite{wgan}.
In the optimization process the discriminator and generator are involved in a min-max game. 
In this game the discriminator aims to maximize the following loss function:

\begin{equation} 
 \begin{split}
    \max_D \underset{{x \sim f_X(x)}}{\mathbb{E}}[D(x|(h,w)]+\underset{{x \sim G(z|(h,w))}}{\mathbb{E}}[-D(x|(h,w)] \\
    +\lambda(||\nabla_{\tilde{x}}  D(\tilde{x}|(h,w)||-1)^2,
    \label{eq:cond_d_loss}
    \end{split}
\end{equation}

where $\lambda$ is a penalty coefficient, $\tilde{x}=\varepsilon x + (1-\varepsilon)G(z|(h,w))$, and  $\varepsilon \sim \text{Uniform}(0,1)$.
The first term is the expected score (which can be thought of as likelihood) of seeing an product $x$ being purchased by the given customer and week $(h,w)$. 
The second term is the score of seeing product $z$ being purchased by that same customer and week, $(h,w)$. 
Taken together, these first two terms encourage the discriminator to maximize the expected score of the real products $x \sim f_X(x)$ given the context $(h,w)$ and minimize the score of the generated products $x \sim G(z|(h,w))$.
The third term in Eq.~\ref{eq:cond_d_loss} is a regularization penalty to ensure that $D$ satisfies the $1$-Lipschitz conditions.

The generator is trained to minimize the following loss function:
\begin{equation} 
    \max_G \underset{{x \sim G(z|(h,w))}}{\mathbb{E}}[D(x|(h,w)]
    \label{eq:cond_g_loss}
\end{equation}
Intuitively, this objective aims to maximize the likelihood that the generated product $x \sim G(z|(h,w))$ is plausible given the context $(h,w)$, where plausibility is determined by the discriminator $D(x|(h,w))$. 
With successive steps of optimization we obtain a $G$ which will generate samples that are more similar to the real data distribution.

While the generator learned from Eq.~\ref{eq:cond_g_loss} can yield realistic product embeddings, in practice one may wish to obtain specific instances from a database $P=\{p_i\}_{i=1}^n$ of known products.
This can be useful, for instance, to obtain a product recommendation for customer $h$ at week $w$.
Given a generated product embedding $G(z|(h,w))$, this can be accomplished by computing the closest product from the database according to the $L_2$ distance metric:
$p=\operatorname{argmin}_{p_i \in P}  ||G(z|(h,w))-p_i||_{2}^{2}$.
Note that other distance metrics such as cosine distance could also be used for this purpose.

\subsection{Generating Sequences of Products} \label{sec:generating_sequences}
In this subsection we develop a pipeline to generate a sequence of baskets of products that a customer will likely purchase over several consecutive weeks.
The pipeline incorporates the product generator $G$ to produce individual products in the basket as well as the LSTM module to model the evolution of a customer's state over a sequence of baskets. 

The procedure works as follows.
Given a new customer with a transaction history $B_1, B_2, \ldots, B_i$, where each $B_i$ denotes a basket for week $w_i$ and $i \geq 1$, we wish to generate a basket $B_{i+1}$ for the following week.
We extract the customer embedding at week $w_i$, denoted $h_i$, by passing the transaction sequence through the LSTM module and extracting the hidden state.
We then find the $k$ most similar customers from the database of known customers by $L_2$ distance from $h_i$.
This is similar to the process of retrieving known products from a database as described in the previous section.
We then determine the number of products to generate for him/her in week $w_i$. To accomplish this we uniformly sample from the basket sizes of the $k$ most similar customers' baskets to retrieve the number of products to generate, $n_i$.
The generator network is then used to generate $n_i$ products via our generator, $G(h_i,w_i)$.

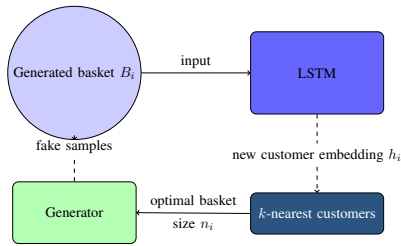
\begin{figure}
\centering
\resizebox{0.3\textwidth}{!}{
\begin{tikzpicture}[node distance=0.8cm][transform canvas={scale=0.5}]

\tikzstyle{LSTM} = [rectangle,rounded corners, minimum width=3cm, minimum height=2cm, text centered,text width=3cm, draw=black, fill=blue!60]
\tikzstyle{Generator} = [rectangle, rounded corners, minimum width=3cm, minimum height=1.5cm, text centered, draw=black, fill=green!30]
\tikzstyle{fake_sample} = [circle, minimum width=1.2cm, minimum height=1.2cm,text centered, draw=black, fill=blue!20]
\tikzstyle{K_nearest2} = [rectangle, rounded corners, minimum width=2cm, minimum height=1cm,text centered,text width=3cm, draw=black, fill={rgb:red,1;green,2;blue,3}]
\tikzstyle{arrow} = [draw, -latex']

\node (main) [Generator]                                                    {Generator};
\node (gg)  [above of=main,yshift=0.8cm]                                    {fake samples};

\node (G_fake) [fake_sample, above of=gg,yshift=1.0cm]                       {Generated basket $B_i$};
\node  (lstm) [LSTM,right of=G_fake, xshift=5.1cm]      {LSTM};
\node (hh)  [below of=lstm,yshift=-1.2cm]                                    {new customer embedding $h_i$};
\node (k_nearest) [K_nearest2, below of=lstm,yshift=-2.62cm]     {\textcolor{white}{$k$-nearest customers}};

\draw[dashed] (main) -- (gg);
\draw[dashed,->]  (gg.north) --  (G_fake.south);


\draw[dashed] (lstm) -- (hh);
\draw[dashed,->] (hh.south) -- (k_nearest.north);
\draw[arrow, ->] (G_fake)  --  (lstm.west) node [above,pos=0.5] {input} node {}   ; 


\draw[arrow,->] (k_nearest.west)  --  (main.east) node  [above,pos=0.5] {optimal basket } node [below,pos=0.5] {size $n_i$}   ; 
\end{tikzpicture}
}
\caption{Basket sequence generation process using the LSTM and Generator modules.}
\label{fig:tikz_seq_gen}
\end{figure}

This procedure can be extended to generate additional baskets by feeding $B_{i+1}$ back into the LSTM, whose hidden state is updated as if the customer had purchased $B_{i+1}$.
The updated customer representation $h_{i+1}$ is once again used to estimate the basket size $n_{i+1}$ and fed into the generator $G(h_{i+1},w_{i+1})$ which yields a basket of products for the week $w_{i+1}$. 
This cycle can be iterated multiple times to generate basket sequences of arbitrary length, or alternatively generate multiple sequences of baskets for the same customer.
The procedure is described in detail by the pseudo-code in Algorithm~\ref{alg:seq_gen} and illustrated in Figure~\ref{fig:tikz_seq_gen}.
Note that all values in Algorithm~\ref{alg:seq_gen} are also indexed by the customer index $c$ which has been omitted in this discussion for the brevity.  To simplify the notation, we also use the symbol $B^c_0$ in Algorithm~\ref{alg:seq_gen} to denote the entire history of customer $c$.

In this manner we can effectively augment a new customer's transaction data by predicting their actions for an arbitrary amount of time.
The intuition behind the approach is that a customer's embedding representation evolves as they purchase products, and therefore might share some common properties with other customers through their purchase experience. 
One can derive insights from this generated data by learning a better characterization of their distribution of likely purchase sequences into the future. 

\begin{algorithm}[h!]
   \caption{Sequence of basket generation}
   \label{alg:seq_gen}
\begin{algorithmic}
   \STATE {\bfseries Input:} LSTM $L$, generator $G$, set of historical basket sequences for each customer $\{B^c_0\}_{c=1}^{C}$, hyperparameter $k$, number of weeks $W$\\
 
 \FOR{$c=1,\ldots,C$}
 \STATE Compute initial customer embedding $h^c_0$ via $L(B^c_0)$
  \FOR{$w=1,\ldots,W$}
   \STATE Sample $n^c_w$ via $k$-nearest customers of $h^c_w$
   \STATE Generate basket $B^c_{w}$ of $n^c_w$ products from $G(h^c_w,w)$
   \STATE Update the customer embedding with the LSTM: $h^c_{w+1} = L(B^c_{w}, h^c_w)$. 
  \ENDFOR
 \ENDFOR
\end{algorithmic}
\end{algorithm}

\section{Experimental Results} \label{sec:experiments}
In this section we empirically demonstrate the effectiveness of the proposed methodology by comparing the generated basket data against real customer data.
Evaluation is first performed with respect to the distributions of key metrics aggregated over the entire data sets, including product categories, brands, prices, and basket sizes. 
Next we compare sequential patterns that exist between products in both data sets, and finally we examine the separability between the real and generated baskets with multiple different basket representations.

\subsection{Experimental Setup}
The basket generation methodology is evaluated using a data set from an industrial partner which consists of 742,686 transactions over a period of 5 weeks during the summer of 2016. This data is composed of 174,301 customer baskets with an average size of $4.08$ products and price of $\$12.2$.
A total of 7,722 distinct products and 66,000 distinct customers exist across all baskets. 

\begin{figure}[t]
    \includegraphics[width=90mm,left]{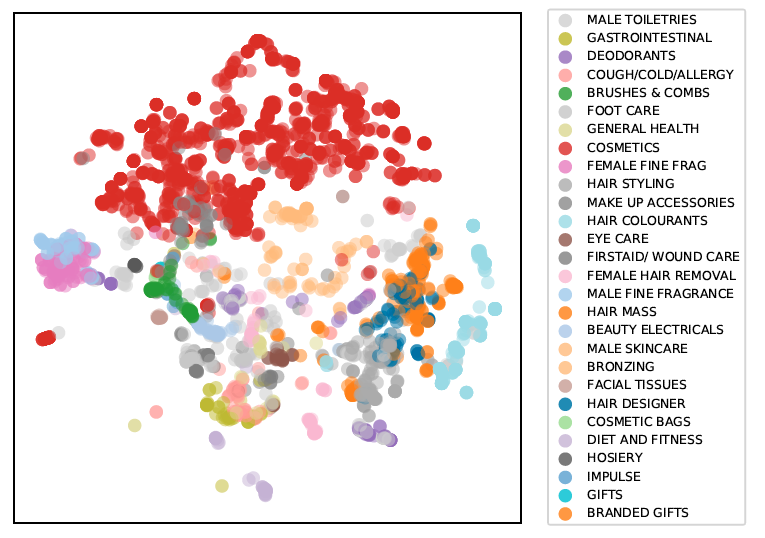}
    \caption{Visualization of product embeddings in a 2D space (mapped using t-SNE)}
    \label{fig:product_tsne}
\end{figure}

Figure \ref{fig:product_tsne} shows the product embedding representations extracted from textual descriptions as described in Section~\ref{sec:product_rep} projected into a 2-dimensional space using the t-SNE algorithm~\cite{tsne}. 
Products are classified into functional categories such as Hair Styling, Eye Care, etc, each of which corresponds to a different color in Figure~\ref{fig:product_tsne}.
We observe that products from the same category tend to be clustered close together, which reflects the semantic relationships between such products. 
At a higher level we observe that similar product categories also occur in close proximity to one another; for example the categories of Hair Mass, Hair Styling and Hair Designer are mapped to adjacent clusters, as are the categories of Female Fine Frag and Male Fine Frag.
This property is critical to the basket generation scheme which directly generates only product embeddings, while instances of specific products are obtained based on their proximity to other products in the embedding space. 

The LSTM is trained on this data set with a multi-task optimization procedure as described in section~\ref{sec:customer_rep} (see Figure~\ref{fig:tikz_multi_task}) for 25 epochs. 
For each customer, we obtain an embedding from the LSTM hidden state after passing through all of their transactions. 
These embeddings are then used to train the conditional GAN. 
The GAN was trained for 100 epochs using the Adam \cite{adam} optimizer with the hyperparameters values of $\alpha=0.5$ and $\beta=0.9$. 
The discriminator is composed of two hidden layers of $256$ units each with ReLU activation functions, with the exception of the last layer which is free of activation functions. 
The generator uses the same architecture except for the last layer which has a tanh activation function. 
During training the discriminator is prioritized by applying five update steps for each update step to the generator. This helps the discriminator converge faster so as to better guide the generator.

Once the LSTM and GAN are trained we run our basket sequence generation. For each customer, we generate $5$ weeks of baskets following the procedure in Algorithm~\ref{alg:seq_gen}.

\subsection{Feature Distributions}
\begin{figure}[h]
    \centering
   \includegraphics[scale=0.4]{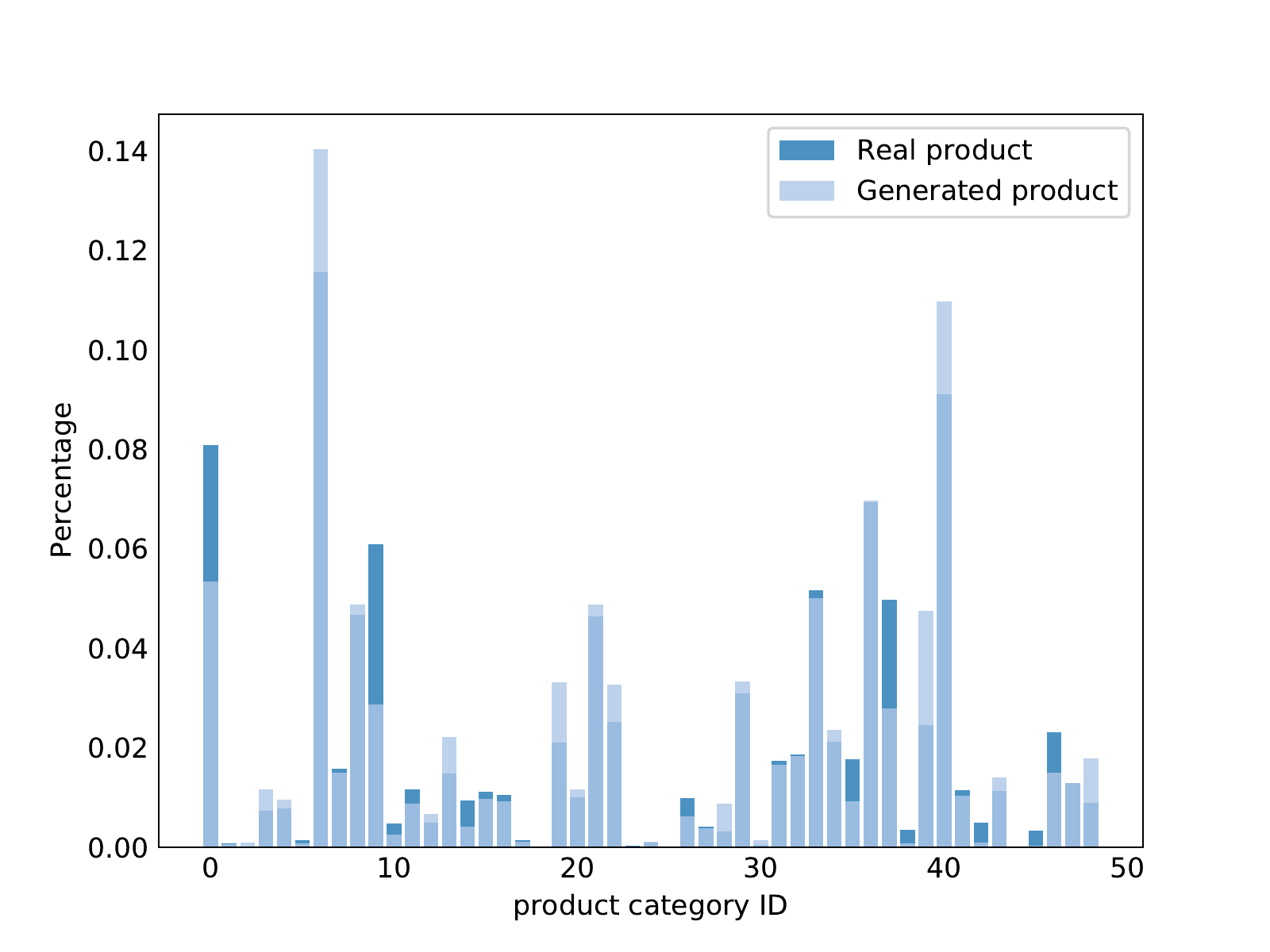}
    \caption{Product category distributions of real and generated data}
    \label{fig:product}
\end{figure}

\begin{figure}[h]
    \centering
   \includegraphics[scale=0.4]{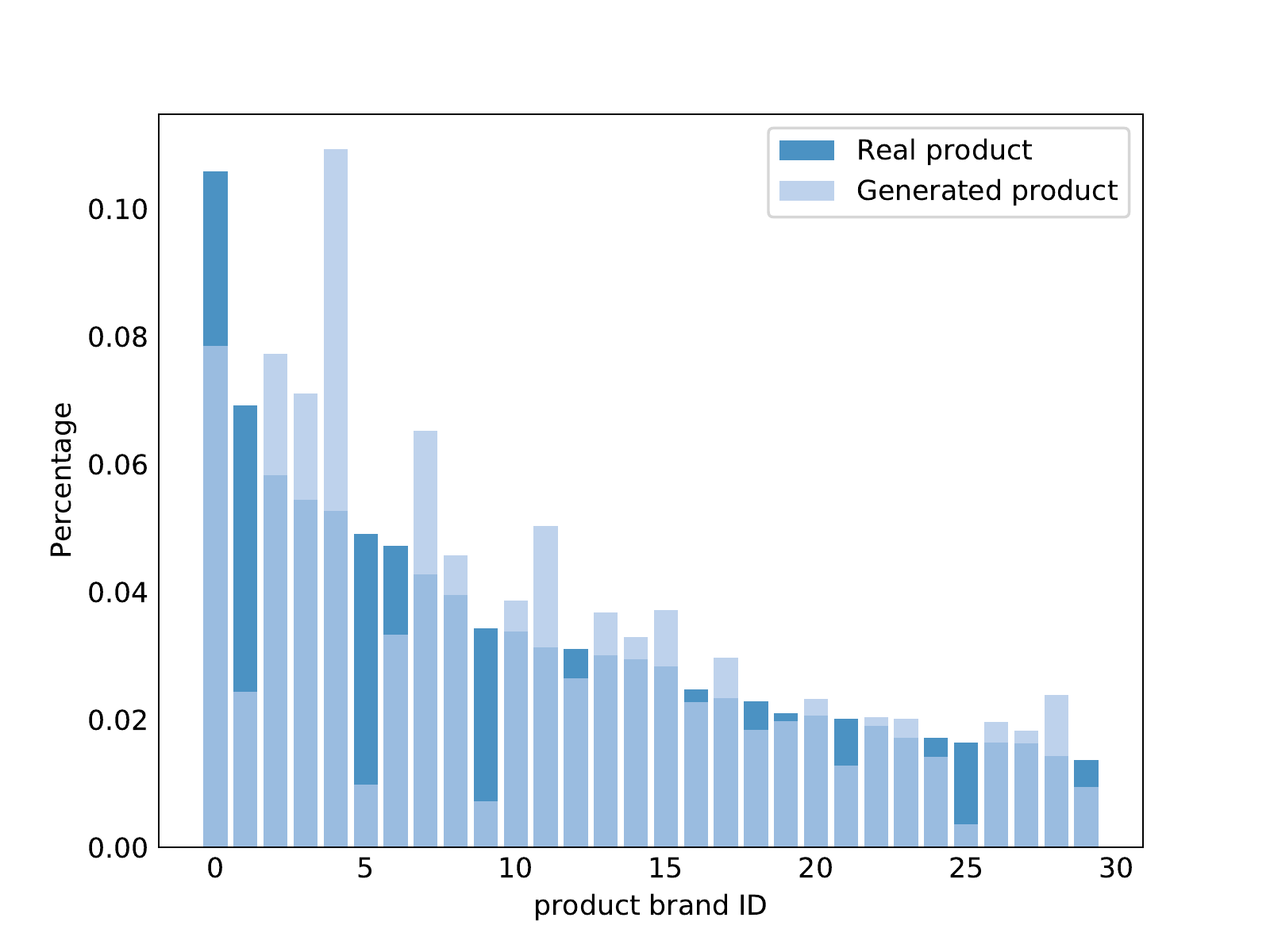}
    \caption{Product brand distributions of real and generated data}
    \label{fig:brand}
\end{figure}

\begin{figure}[h]
    \centering
   \includegraphics[scale=0.4]{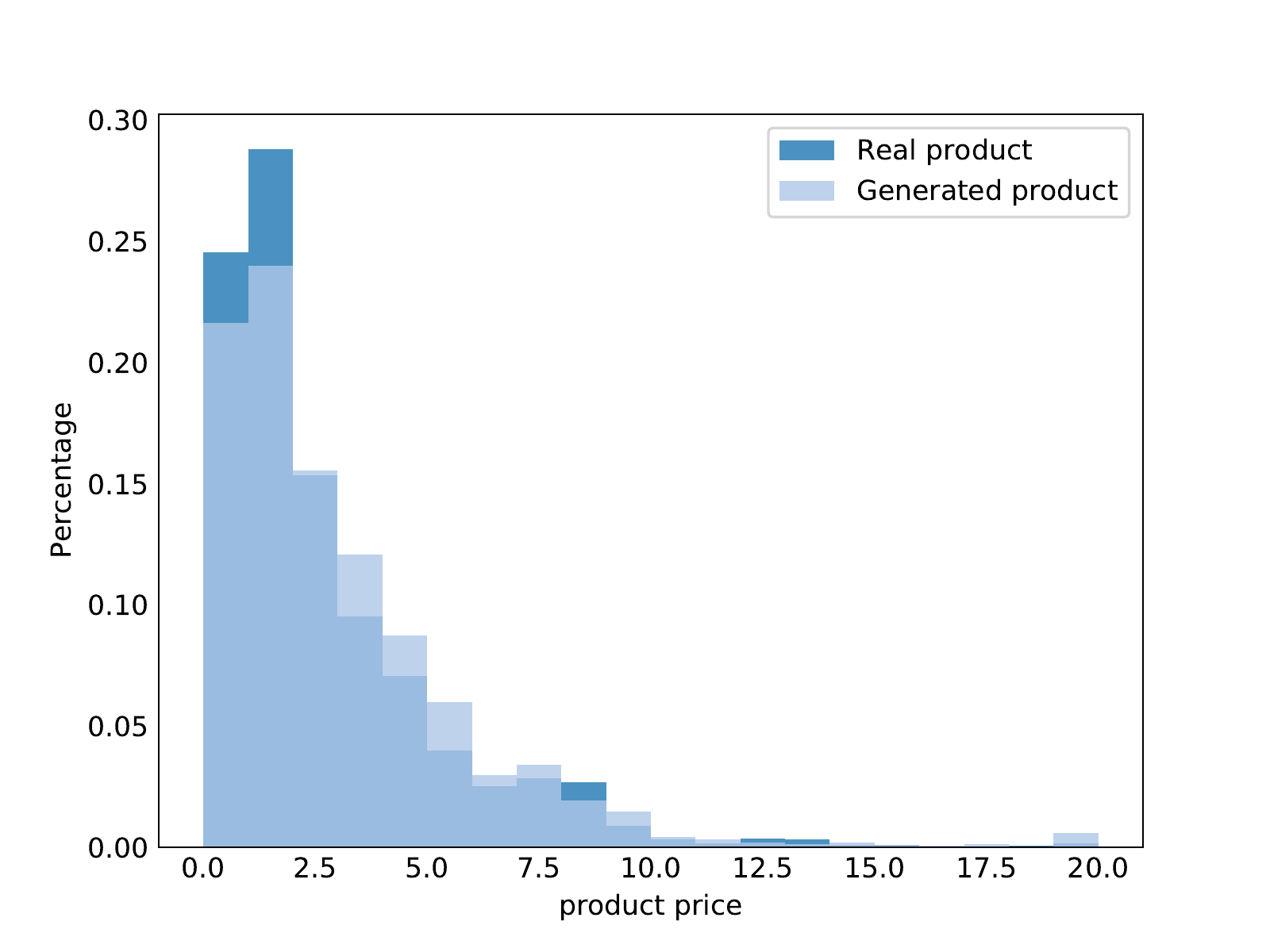}
    \caption{Product price distributions of real and generated data}
    \label{fig:price}
\end{figure}

\begin{figure}[h]
    \centering
   \includegraphics[scale=0.4]{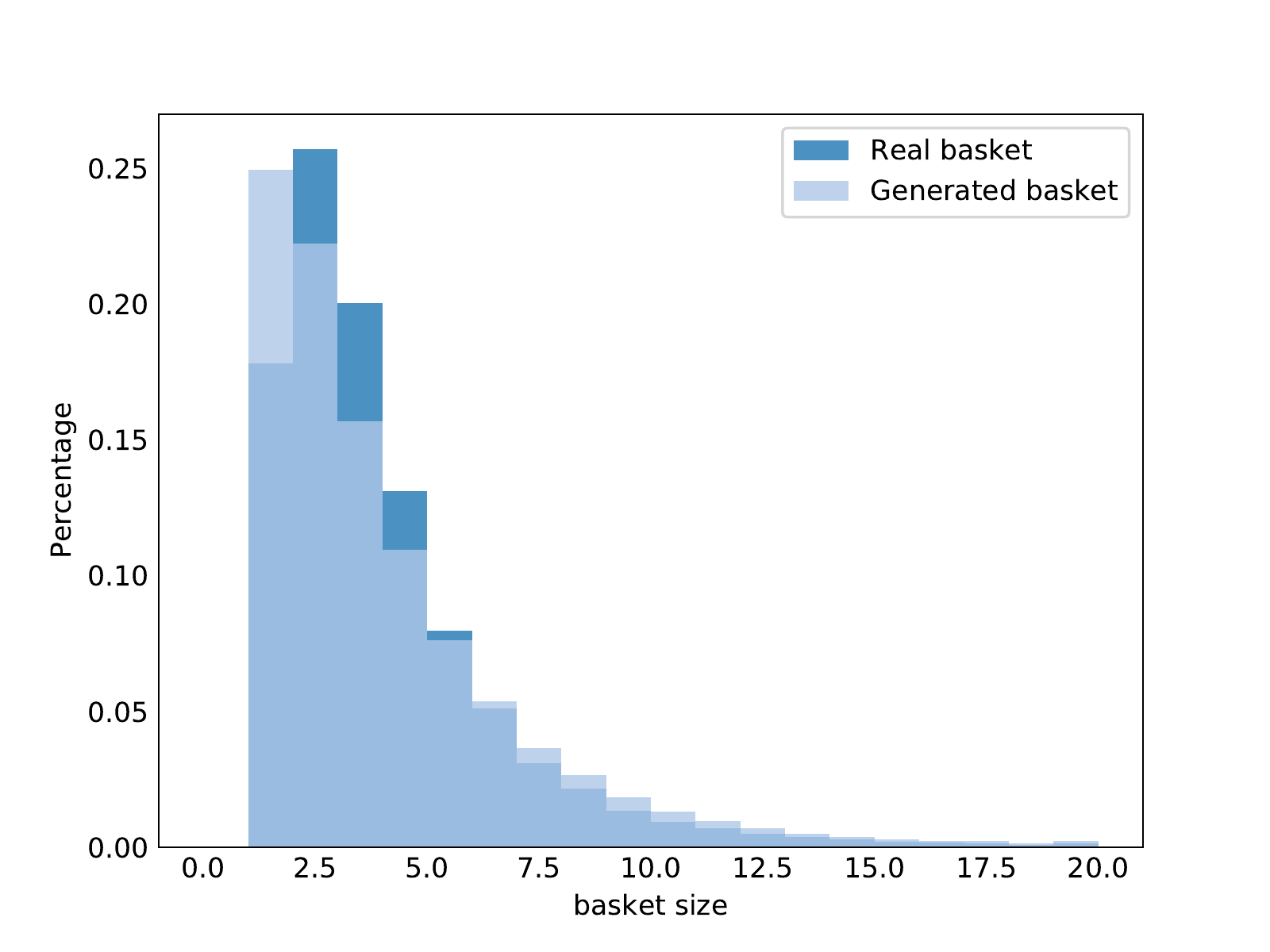}
    \caption{Basket size distributions of real and generated data}
    \label{fig:basket_size}
\end{figure}

Figures \ref{fig:product}, \ref{fig:brand}, \ref{fig:price}, and \ref{fig:basket_size} compare the frequency distributions of the categories, brand, prices, and basket sizes, respectively, between the generated and real baskets. Additional metrics are provided in Tables \ref{fig:statistics} and \ref{fig:discrpancies}. Note that for the brand, we restrict the histogram plots to include only the top $30$ most frequent brands.
We observe that in general our generative model can reasonably replicate the ground-truth distribution. This is further evidenced by Table \ref{fig:discrpancies} which indicates that the highest absolute difference in frequency of generated brands is $5.6\%$. 
The lowest discrepancy occurs for the category feature, where the maximum deviation is $3.2\%$ in the generated products. 
In addition, the generated basket size averages $3.85$ products versus $4.08$ for the real data which is a difference of approximately $5\%$. 
The generated product prices are an average of $\$3.1$ versus $\$3.4$ for the real data (a $10\%$ difference). 
This demonstrates that the generation methodology can mimic the aggregated statistics of the real data to a reasonable degree.  Note that we should not expect the two distributions to match exactly because we are projecting each customer's purchases into the future, which won't necessarily have the same distributive properties.

\begin{table}[ht]
\caption{Statistics}
\centering
\resizebox{0.5\textwidth}{!}{
\begin{tabular}{|l|c|c|c|}
\hline
          &  Real Transactions     &   Generated Transactions    \\     \hline 
Average basket size  	   &      $4.08$       &    $3.85$             	\\  \hline   
Average basket price    	   	   &   $\$3.1$          &    $\$3.4$               	\\  \hline   
\end{tabular}}
\label{fig:statistics}
\end{table}

\begin{table}[ht]
\caption{Discrepancies between real and generated data}
\centering
\resizebox{0.4\textwidth}{!}{
\begin{tabular}{|l|c|c|}
\hline
Criterion          &  Max absolute deviation (in $\%$)\\     \hline 
Category  	   &      $3.2\% $               	\\  \hline   
Brand    	   	   &   $5.6\% $           	\\  \hline 
Price    	   	   &   $5.2\% $           	\\  \hline  
Basket size*  	   	   &   $4.1\% $           	\\  \hline  
\end{tabular}
}
\begin{tablenotes}
\item $^*$ this metric only applies for basket size $\leq 20$
\end{tablenotes}

\label{fig:discrpancies}
\end{table}

\begin{table*}[ht]
\caption{Sequential patterns comparison between real and generated transaction data}
\centering{
\resizebox{0.9\textwidth}{!}{
\begin{tabular}{|l|c|c|}
\hline
 Sequence &  Real support & Generated support \\
 \Xhline{3\arrayrulewidth}
Hemorrhoid relief, Skin treatment \& dressings & 0.045 & 0.098 \\ \hline
Skin treatment \& dressings, Female fine frag & 0.029 & 0.100 \\ \hline
Facial moisturizers, Skin treatment \& dressings & 0.028 & 0.075 \\ \hline
Shower products, Female fine frag & 0.028 & 0.056 \\ \hline
Hemorrhoid relief, Female fine frag & 0.028 & 0.093 \\ \hline
Skin treatment \& dressings, Facial moisturizers & 0.027 & 0.076 \\ \hline
Skin treatment \& dressings, Preg test \& ovulation & 0.027 & 0.082 \\ \hline
Shower products, Skin treatment \& dressings & 0.026 & 0.056 \\ \hline
Hemorrhoid relief, Preg test \& ovulation & 0.026 & 0.075 \\ \hline
Female fine frag, Preg test \& ovulation & 0.025 & 0.081 \\ \hline
Facial moisturizers, Hemorrhoid relief & 0.025 & 0.069 \\ \hline
Skin treatment \& dressings, Skin treatment \& dressings, Hemorrhoid relief & 0.007 & 0.014 \\ \hline
\end{tabular}}
}
\label{tab:association_rule}
\end{table*}

\subsection{Sequential Pattern Mining}

Sequential pattern mining~\cite{pattern_mining} (SPM) is a technique to discover
statistically relevant subsequences from a sequence of sets ordered by time.
One frequent application of SPM is in retail transactions where we wish to determine subsequences of
items across baskets customers have bought over time.  For example, given
an set of baskets a customer has purchased ordered by time: 
\{{\em milk, bread}\}, \{{\em cereal, cheese}\}, \{{\em bread, oatmeal, butter}\},
one sequential pattern we can derive is: \{{\em milk}\}, \{{\em bread, butter}\}
because \{{\em milk}\} in the first basket comes before \{{\em bread, butter}\}
in the last basket.
A pattern is typically measured by its {\em support}, which is defined as the
number of customers containing the pattern as a subsequence.
We refer the reader to~\cite{pattern_mining} for further details.

\begin{figure}[ht]
    \centering
   \includegraphics[width=90mm]{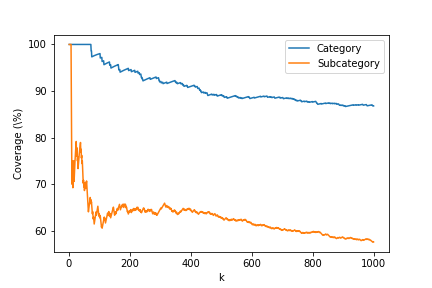}
    \caption{Generated pattern coverage of the top-$k$ most common real patterns}
    \label{fig:coverage}
\end{figure}

For this set of experiments, sequential pattern mining is performed on the real
and generated datasets via the SPFM~\cite{spmf_seq_mining} library using a
minimum support of 1\% of the total number of customers.
Figure~\ref{fig:coverage} plots the
percentage of the top-$k$ most common real sequential patterns that are also found in the generated data as $k$ varies from 1 to 1000. 
Here items are defined at either the category or subcategory level, so that two products are considered equivalent if they belong to the same functional
grouping. We see that for the category-level, we can recover 98\%
of the top-100 patterns, while at the subcategory-level, we can recover 63\%.  
This demonstrates that our method is generating plausible
sequences of baskets for customers because most of the real sequential patterns
show up in the generated data.  Not all patterns are found however, which might
imply that the generated data might have some drift in the sequences due to the
method of projecting customer's purchases into the future.

Table \ref{tab:association_rule} shows examples of the top sequential patterns of length 2 and 3 from the real data at the subcategory level that also appeared
in the generated transactional data.  The two right columns show the support
for both the real and generated datasets, which is normalized by dividing by the
total number of customers.  We can see that the generated data has higher
support for the patterns from generated data, indicating that it may have an easier time replicating common patterns.

\subsection{Basket Distributions}
In this section we directly compare the generated and real baskets based on the products they contain.
For each basket of products $B_i=\{p_{i,j}\}_{j=1}^{|B_i|}$ we a define a vector representation $v_i$ using a bag-of-products scheme. 
Let $P$ denote the set of all known products. 
Then $v_i$ is a $|P|$-dimensional vector with $v_i^{(j)}=1$ if $p_j \in B_i$ or $v_i^{(j)}=0$ otherwise.
$P$ can be defined at various levels of precision such as the product serial number, the brand, or the category levels. 
At the category level, for instance, two products would be considered equivalent and correspond to the same index $j$ if they belong to the same category.

\begin{figure}[ht]
    \centering
   \includegraphics[width=90mm]{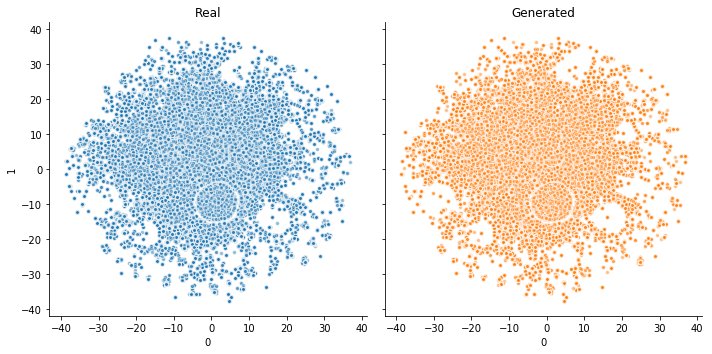}
    \caption{Basket representations as bags-of-products vectors at the category level, projected using t-SNE.}
    \label{fig:boi_category_tsne}
\end{figure}

\begin{figure}[ht]
    \centering
   \includegraphics[width=90mm]{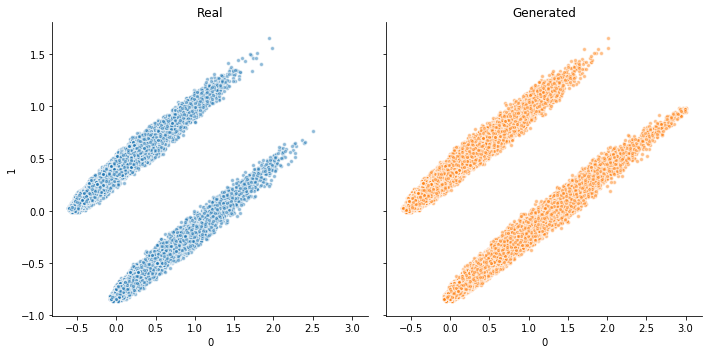}
    \caption{Basket representations as bags-of-products vectors at the category level, projected using PCA.}
    \label{fig:boi_category_pca}
\end{figure}

The resulting vectors are then projected into two dimensions using t-SNE for visualization purposes. 
The distributions of the real and generated data are plotted in Figures~\ref{fig:boi_category_tsne}.
For an alternative viewpoint Figure~\ref{fig:boi_category_pca} plots the vectors projected using Principal Component Analysis (PCA). 
These plots qualitatively indicate that the distributions match quite closely.

This observation can be further analyzed quantitatively by training a classifier to distinguish between points from the two distributions.
By measuring the prediction accuracy of this classifier we obtain an estimate of the degree of separability between the data sets.
For our experiments we randomly sample a subset of the generated points such that the number of real and generated points are equal.
This way a perfectly indistinguishable generated data set should yield a classification accuracy of 50\%. 
We note that this classification task is fundamentally unlike that which is performed by the discriminator during the GAN training, as the latter operates on the embedding representation of a single product while the former operates on the bag-of-items representation of a basket. 

The results are given in Table~\ref{tab:basket_sep_acc} using a logistic regression classifier. 
Each row corresponds to a different level of granularity in the definition of the bag-of-products representation, with category being the most coarse--grained and sku being the most fine--grained.
We see that the classifier performs quite poorly at the category levels, meaning that the generated baskets of categories are quite plausible.

However, note that the bag-of-products representation does not preserve the semantic similarity between products in that any two products with different skus are perfectly separable even if they have very similar functions and descriptions.
Therefore, we instead define the sku level basket representation as the mean of embeddings of the products in the basket. 
This is given in the last row of Table~\ref{tab:basket_sep_acc}.
Note that these representations come from the embeddings of the nearest neighbor product rather than the output of the generator. 
As expected, the classification accuracy is still quite low considering the fine-grained level at which the prediction occurs.

\begin{table}[H]
    \caption{Separability between real and generated baskets.}
    \centering
    \resizebox{0.4\textwidth}{!}{
    \begin{tabular}{|c|c|}
        \hline
        Basket Representation & Classification Accuracy \\ 
         \Xhline{2\arrayrulewidth}
        Bag-of-products category & 0.634 \\ \hline
        Bag-of-products subcategory & 0.663 \\ \hline
        Basket embedding sku-level & 0.704 \\ \hline
    \end{tabular}
    }
    \label{tab:basket_sep_acc}
\end{table}




\section{Conclusion}

In this paper, we propose a novel method of generating sequences of realistic customer baskets for customer-level transactional data. After learning customer embeddings with an LSTM, we generate a product basket conditioned on the customer embedding using the generator from the GAN.  The generated basket of products is fed back into the LSTM to generate a new customer embedding, and the process repeats. We show that the proposed methods can replicate to a reasonable degree the statistics of the real data distribution (category, brand, price and basket size). As additional experiments, we verified that common sequential patterns exist between products in the generated and real data, and that the generated orders are difficult to distinguish from the real orders.

\newpage
\bibliographystyle{unsrt}

\end{document}